\documentclass[letterpaper]{article} % DO NOT CHANGE THIS
\usepackage{aaai24}  % DO NOT CHANGE THIS
\usepackage{times}  % DO NOT CHANGE THIS
\usepackage{helvet}  % DO NOT CHANGE THIS
\usepackage{courier}  % DO NOT CHANGE THIS
\usepackage[hyphens]{url}  % DO NOT CHANGE THIS
\usepackage{graphicx} % DO NOT CHANGE THIS
\usepackage{natbib}  % DO NOT CHANGE THIS AND DO NOT ADD ANY OPTIONS TO IT
\usepackage{caption} % DO NOT CHANGE THIS AND DO NOT ADD ANY OPTIONS TO IT
\usepackage{algorithm}
\usepackage{algorithmic}

\usepackage{amsmath}
\usepackage{footmisc}
\usepackage{color,xcolor}
\usepackage{bm}
\usepackage{amssymb}
\usepackage{amsthm}

\theoremstyle{plain}
\newtheorem{theorem}{Theorem}

\newtheorem{lemma}{Lemma}

\theoremstyle{definition}

\theoremstyle{remark}

\makeatletter
\renewcommand{\@makefnmark}{\hbox{\@textsuperscript{\normalfont\@thefnmark}}}
\makeatother

\usepackage{newfloat}
\usepackage{listings}
\DeclareCaptionStyle{ruled}{labelfont=normalfont,labelsep=colon,strut=off} % DO NOT CHANGE THIS
\floatstyle{ruled}
\newfloat{listing}{tb}{lst}{}
\floatname{listing}{Listing}
\title{Explaining Generalization Power of a DNN Using Interactive Concepts}
\author{
    Huilin Zhou\textsuperscript{\rm 1},
    Hao Zhang\textsuperscript{\rm 1},
    Huiqi Deng\textsuperscript{\rm 1},
    Dongrui Liu\textsuperscript{\rm 1},\\
    Wen Shen\textsuperscript{\rm 1},
    Shih-Han Chan\textsuperscript{\rm 1, 2},
    Quanshi Zhang\textsuperscript{\rm 1}\thanks{Quanshi Zhang is the corresponding author. He is with the Department of Computer Science and Engineering, the John Hopcroft Center, at the Shanghai Jiao Tong University, China. Correspondence to: Quanshi Zhang $<$ zqs1022@sjtu.edu.cn$>$.}
}
\affiliations{
    \textsuperscript{\rm 1}Shanghai Jiao Tong University\\
    \textsuperscript{\rm 2}University of California San Diego\\
    $\{$zhouhuilin116, 1603023-zh, denghq7, drliu96, wen\_shen$\}$@sjtu.edu.cn\\
    s2chan@ucsd.edu, zqs1022@sjtu.edu.cn
    
}

\usepackage{bibentry}
\begin{document}

\maketitle

\begin{abstract}
This paper explains the generalization power of a deep neural network (DNN) from the perspective of interactions.
Although there is no universally accepted definition of the concepts encoded by a DNN, the sparsity of interactions in a DNN has been proved, \emph{i.e.}, the output score of a DNN can be well explained by a small number of interactions between input variables. In this way, to some extent, we can consider such interactions as interactive concepts encoded by the DNN.
Therefore, in this paper, we derive an analytic explanation of inconsistency of concepts of different complexities. This may shed new lights on using the generalization power of concepts to explain the generalization power of the entire DNN.
Besides, we discover that the DNN with stronger generalization power usually learns simple concepts more quickly and encodes fewer complex concepts. 
We also discover the detouring dynamics of learning complex concepts, which explains both the high learning difficulty and the low generalization power of complex concepts. The code will be released when the paper is accepted.
\end{abstract}

%%%%%%%%% BODY TEXT
\section{Introduction}
\label{sec:intro}

Although deep neural networks (DNNs) have achieved remarkable success 
%in various tasks 
nowadays, the essence for the superior generalization power of a DNN is still unclear. People usually explained DNNs via the flatness of the loss landscape~\cite{flat_minima} and theoretical bounds for the generalization~\cite{pac_bound, norm_bound}, or by proposing new metrics for the representation power~\cite{stiffness, CLEVER}. In recent years, analyzing the capacity or blind spots of encoding specific concepts represents an emerging direction in explaining DNNs~\cite{deng2021bottleneck, chengxv}.

Therefore, unlike previous studies, we revisit the generalization of a DNN from a new perspective of concepts. If the inference score of a DNN can be attributed to a set of countable concepts, then the generalization power of a DNN would be explained by the generalization power of elementary concepts encoded by the DNN. This will be a new insight into the generalization power of the DNN.

\begin{figure}[t]
   \centering
   \includegraphics[width=0.9\linewidth]{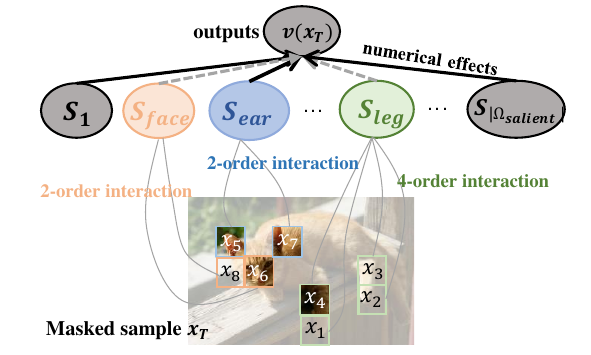}
   \caption{Interactions encoded by the DNN. Each interaction $S$ represents an AND relationship between a set of input variables (\emph{e.g.}, image regions). 
   Masking any patches in $S_{\text{face}}$ will deactivate the interaction, making $I(S_{\text{face}} \vert \boldsymbol{x})=0$.}
   \label{fig:causal_graph}
\end{figure}

\textbf{Can we really define concepts encoded by a DNN?} 
How to define concepts encoded by a DNN is still an open problem, and there is no formal and universally-accepted definition so far. To this end, given a trained DNN, \citet{ren2021AOG} quantified Harsanyi interactions~\cite{harsanyi} encoded by a DNN as concepts in the DNN. 
Each Harsanyi interaction represents an AND relationship between a set of input variables. For example, in the natural language processing, a Harsanyi interaction may represent the AND relationship between words. In the face detection, as Figure~\ref{fig:causal_graph} shows, a Harsanyi interaction may consist of patches of {\small $S=\{ \text{eyes}, \text{nose}, \text{mouth}\}$} to represent the AND relationship between image patches for the face. Only if all patches in {\small $S$} are present in the image, the face interaction {\small $S$} is activated and makes a numerical effect {\small $I(S \vert \boldsymbol{x})$} on the detection score. 

Although \citet{ren2021AOG} did not convince us that the above interaction really represented a concept that fits human cognition, they did provide mathematical supports for such interactions. 
(1) \citet{li2023does, ren2021AOG} discovered and~\citet{ren2023arrived} mathematically proved that a well-trained DNN usually encoded just a few interactions between different input variables. 
(2) We can use these interactions to explain the output of a DNN. 
(3) Besides, ~\citet{li2023does} found the high transferability of these interactions across samples and across models.

Above findings partially guarantee that Harsanyi interactions can be roughly taken as sparse primitives responsible for the network output. 
Thus, \textbf{these interactions provide us a more straightforward way to redefine the representation power of a DNN.} 
As in~\cite{ren2021AOG}, let us just call such sparse interactions as \textbf{\textit{interactive concepts}} encoded by the DNN.
Let an interactive concept be frequently extracted by the DNN from training samples, \emph{e.g.}, a common face concept {\small $S = \{ \text{eyes}, \text{nose}, \text{mouth} \}$} shared by different images.
\textbf{If this concept also frequently appears in testing samples, then this concept is considered generalizable; otherwise, not generalizable.} Because the network output is proved to be the sum of effects of different interactive concepts, the generalization of concepts can be a deep insight into the generalization of the entire DNN.
Consequently, an out-of-the-distribution (OOD) sample will be explained to contain some non-generalizable interactive concepts\footnote{Please see Section 1 in supplemental materials for details.}.

Although there is a common intuition that more complex representations usually lead to over-fitting, this study uses an analytic inconsistency of concepts to explain the connection between the complexity of interactive concepts and their generalization power. The complexity of an interactive concept {\small $S$} is defined as the number of input variables in {\small $S$}, which is also termed as the \textit{order} of the concept, \emph{i.e.}, {\small $\text{order}(S)=\vert S\vert$}.
Therefore, a high-order interactive concept contains a large number of input variables, and represents a complex concept.
In this way, we use the high inconsistency to noises of high-order concepts to explain the high over-fitting risk of high-order concepts. 

Besides, the high over-fitting risk of high-order concepts can also be explained by the detouring dynamic of learning high-order concepts. \emph{I.e.}, we find that a high-order concept is more likely to be mistakenly represented by the DNN as a mixture of low-order concepts. We also find the following four phenomena to explain the high over-fitting risk of high-order concepts.

$\bullet$ For each concept, we compute the distribution of its effects on training samples and such a distribution on testing samples. We find that compared to the distribution of high-order concepts, the distribution of low-order concepts in training samples and that in testing samples are usually more similar to each other. This indicates the strong generalization power of low-order concepts.

$\bullet$ Under adversarial perturbations, high-order concepts are more likely to make inconsistent interaction effects than low-order concepts. 

$\bullet$ Let us focus on a set of DNNs with the same architecture, which are trained at different over-fitting levels. We find that over-fitted DNNs usually encode stronger high-order interactive concepts than normal DNNs.

$\bullet$ Besides, normal DNNs usually learn low-order interactive concepts faster than over-fitted DNNs.
% 实验中 verifiable prediction of 。。。。？

\textbf{Interactive concepts vs. cognitive concepts and other interaction metrics.} Although the Harsanyi interactive concept seems partially aligned with humans’ cognition to some extent~\cite{cheng2021proto}, we do not think such interactive concepts exactly fit humans’ cognition. More crucially, the mathematical generalization power of a concept (defined in Equation~\eqref{eq:similarity}) does not depend on whether the concept fits human cognition. To this end, \citet{ren2021AOG} have proved that the Harsanyi interaction could represent primitives of inference logic of a DNN, which was already sufficient for our research. Please see Section 2 in supplemental materials for detailed comparisons between the Harsanyi interaction and other interaction metrics.

In general, a DNN's representation complexity is different from the cognitive complexity. For example, let us consider a small ball concept consisting of a few pixels (low-order concept) and a large ball concept consisting of massive pixels (high-order concept) in images. These two balls have similar cognitive difficulty. However, from the perspective of a DNN, a large ball has more pixels, so that the DNN has to examine whether all pixels within the large ball share the same color without exceptions. This is more difficult than examining a few pixels within a small ball.

% \textcolor{blue}{\textbf{Practical values and connections to existing findings.} \citet{ren2023bayesian} have proved that high-order interactive concepts are
% more difficult to be learned by DNNs. \citet{deng2021bottleneck} proposed two dedicated loss functions to encourage and penalize the DNN to encode specific order interactions for inference. Our research may provide conceptual and analytic insights into the significance of these previous studies.}

\section{Explaining Generalization Using Concepts}
\label{sec:algorithm}
Currently, there is no formal and universally-accepted definition for concepts encoded by DNNs.
In this paper, we follow \citet{ren2021AOG} to take the Harsanyi interaction as a simplified definition of concepts or primitives encoded by a DNN. These interactions are proved to well mimic network outputs under different input variations, so we can roughly consider such concepts as primitives to analyze the DNN. Our analysis does not require the exact fitness between the concept and human cognition.

%-------------------------------------------------------------------------
\subsection{Preliminaries: Sparse Interactive Concepts}
\label{sec:interactive concepts}
\citet{li2023does},~\citet{ren2021AOG} discovered and~\citet{ren2023arrived} mathematically proved that \textbf{a well-trained DNN usually only encoded a small number of interactions between input variables}.
Specifically, given a well-trained DNN {\small $v$} and an input sample {\small $\boldsymbol{x}=[x_1, x_2, \ldots, x_n]^{\top}$} with {\small $n$} input variables, let {\small $N=\{ 1, 2, \ldots, n \}$} denote the indices of all {\small $n$} input variables in {\small $\boldsymbol{x}$}, and let {\small $v(\boldsymbol{x}) \in \mathbb{R}$} denote the scalar output of the DNN or a certain output dimension of the DNN\footnote{Note that people can apply different settings for {\small $v(\boldsymbol{x})$}. In particular, for multi-category classification tasks, we set {\small $v(\boldsymbol{x}) = \text{log} \frac{p(y=y^{\text{truth}} \vert \boldsymbol{x})}{1 - p(y=y^{\text{truth}} \vert \boldsymbol{x})} \in \mathbb{R}$} by following~\cite{deng2021bottleneck}.}. 
Then, the Harsanyi dividend (or Harsanyi interaction)~\cite{harsanyi} is used to quantify the effect of the interaction between a set {\small $S\subseteq N$} of input variables. 
\begin{equation}
\small
\forall S \subseteq N, I(S \vert \boldsymbol{x}) = \sum\nolimits_{T \subseteq S} (-1)^{\vert S \vert - \vert T \vert} \cdot v(\boldsymbol{x}_T),
\label{eq:harsanyi_interaction}
\end{equation}
where {\small $\boldsymbol{x}_T$} represents a sample whose input variables in {\small $N \setminus T$} are masked by baseline values\footnote{The baseline value of each input variable is usually implemented as the mean value of this input variable over all samples~\cite{baseline_mean}.}. 

Each interaction with considerable effect {\small $I(S \vert \boldsymbol{x})$} represents the AND relationship between input variables in {\small $S$}.
For example, in Figure~\ref{fig:causal_graph}, the face interaction consists of image patches in the set {\small $S_{\text{face}}=\{ \text{eyes}, \text{mouth},  \text{nose}\}$}. Only when all three image patches in {\small $S_{\text{face}}$} are present in the input sample {\small $\boldsymbol{x}$}, the face interaction {\small $S_{\text{face}}$} is activated, and makes a numerical effect {\small $I(S_{\text{face}}\vert \boldsymbol{x})$} on the network output. Otherwise, if any image patch in {\small $S_{\text{face}}$} is masked in the input image {\small $\boldsymbol{x}$}, then we can no longer measure a numerical effect of the interaction, \emph{i.e.}, getting {\small $I(S_{\text{face}}\vert \boldsymbol{x}_{\text{masked}})=0$} based on Equation \eqref{eq:harsanyi_interaction}.

%the face concept {\small $S_{\text{face}}$} is deactivated, \emph{i.e.}, making {\small $I(S_{\text{face}}\vert \boldsymbol{x})=0$}.

% merge
\textbf{Sparsity \& universal matching.}
\quad \citet{ren2023arrived} have proved that although there are {\small $2^n$} different combinations of variables $S \subseteq N$, as Figure~\ref{fig:sparsity} shows, 
most well-trained DNNs\footnote{Please see Section 4 in supplemental materials for detailed common conditions for the emergence of sparse interactions.} only encode a small number of interactions (combinations) {\small $S \in \Omega_{\text{salient}}$} with salient effects {\small $I(S \vert \boldsymbol{x})$}, subject to {\small $\vert \Omega_{\text{salient}} \vert \ll 2^n$}. All other interactions measured in Equation \eqref{eq:harsanyi_interaction} have almost zero effects, {\small $I(S \vert \boldsymbol{x}) \approx 0$}, which represent noisy patterns.

\begin{theorem}[]
An input sample {\small $\boldsymbol{x}$} can be masked in {\small $2^n$} ways by sampling different {\small $T \subseteq N$}. For any randomly masked sample {\small $\boldsymbol{x}_T$}, ~\citet{ren2021AOG} have proved that%the output score of the DNN {\small $v(\boldsymbol{x}_T)$} can be approximated by the sum of effects of a small number of salient interactions.
\begin{equation}
\small
    v(\boldsymbol{x}_T) = \sum\nolimits_{S \subseteq T}I(S\vert \boldsymbol{x}) \approx \sum\nolimits_{S\subseteq T : S \in \Omega_{\text{salient}}}I(S\vert \boldsymbol{x})
\label{eq:faithfulness}
\end{equation}
\label{th:faithfulness}
\end{theorem}
Based on the proved sparsity, Theorem \ref{th:faithfulness} further indicates that network outputs on all {\small $2^n$} randomly masked samples {\small $\{ \boldsymbol{x}_T : T \subseteq N \}$} can be \textbf{universally approximated} by a small number of salient interactions in {\small $\Omega_{\text{salient}}$}, subject to {\small $\vert \Omega_{\text{salient}} \vert \ll 2^n$}. According to Occam's Razor~\cite{occam}, if the inference score can be explained as just a small number of concepts, then the concept is more likely to reflect the essential knowledge encoded by a DNN, instead of a mathematical trick without clear meanings. In this way, interactive concepts can be defined as follows.

\textbf{Definition of interactive concepts.} \quad Considering the proved sparsity, an interactive concept is defined as a salient interaction. Given a threshold {\small $\tau$}, the set of interactive concepts are defined as {\small $\Omega_{\text{salient}} = \{ S \subseteq N: \vert I(S \vert \boldsymbol{x}) \vert > \tau \}$}. 

\begin{figure*}[t]
  \centering
  \includegraphics[width=0.85\linewidth]{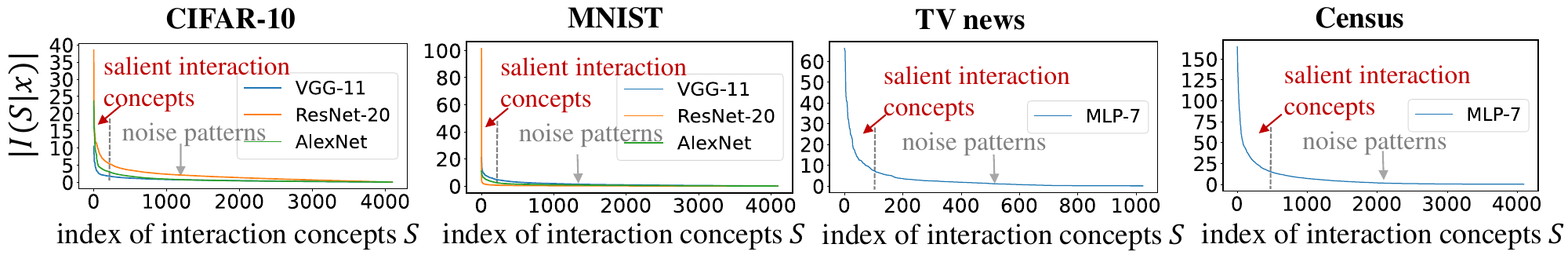}
  \caption{Interactive concepts encoded by a DNN are usually very sparse. This phenomenon exists in various DNNs trained on different datasets. We sort the interactive concepts to a decreasing order of the interaction strength {\small $\vert I(S \vert \boldsymbol{x}) \vert$}.}
  \label{fig:sparsity}
\end{figure*}

% \begin{figure*}[t]
%   \centering
%   \begin{minipage}{0.8\linewidth}
%       \centering
%       \includegraphics[width=\linewidth]{figures/sparsity.pdf}
%   \end{minipage}%
%   \begin{minipage}{0.2\linewidth}
%       \caption{Interactive concepts encoded by a DNN are usually very sparse. We sort the concepts to a decreasing order of the strength {\small $\vert I(S \vert \boldsymbol{x}) \vert$}.}
%       \label{fig:sparsity}
%   \end{minipage}
% \end{figure*}

\textbf{Mathematical and experimental supports for interactive concepts.}
\quad (1) First, although there is no theory to ensure such a simplified definition exactly fits concepts in human cognition, it is proved that this definition mathematically guarantees that the output of DNNs can be approximated by sparse interactive concepts.
(2) Besides,~\citet{li2023does} observed that interactive concepts also had certain \textbf{transferability} across samples and across models, \emph{i.e.}, concepts in one sample could also appear in another sample in the same category, and concepts encoded by a DNN were usually also encoded by other DNNs. 
(3) Finally, a salient interactive concept also exhibited \textbf{strong discrimination power}, \emph{i.e.}, if a set of samples all have the same salient concept, then this concept will probably push these samples towards the same category in classification.
% 迁移性（展开一句，在不同的。。。都。。。相似的。。。。），分类性（展开一句）

% Meanwhile, Ren et al.~\cite{ren2023arrived} has proven that such \textbf{sparse salient concepts} could \textbf{universally approximate} network outputs of all $2^n$  randomly masked samples $\forall T \subseteq N, \boldsymbol{x}_T$.
% \begin{equation}
% \forall T \subseteq N, v(\boldsymbol{x}_T) = \sum\nolimits_{S \subseteq T}I(S\vert \boldsymbol{x}) \approx \sum\nolimits_{S\subseteq T : S \in \Omega_{\text{salient}}}I(S\vert \boldsymbol{x})
% \label{eq:sum_of_I_salient}
% \end{equation}
% According to Occam's Razor~\cite{occam}, if the inference score can be explained as just a small number of concepts, rather than a large number of concepts, then the concept is more likely to reflect the essential knowledge encoded by a DNN, instead of a mathematical trick that satisfies Equation~(\ref{eq:network_output}).

\textbf{Complexity (order) of interactive concepts.}
\quad The complexity of the interactive concept $S$ is defined as the number of input variables contained in the concept, which is also termed as the \textbf{\textit{order}} of the concept, \emph{i.e.}, {\small $\textit{order}(S)=\vert S \vert$}.
A low-order (simple) interactive concept represents interactions between a small number of input variables. A high-order (complex) interactive concept represents a complex interaction between a large number of input variables.

% \textbf{Modeling complexity vs. cognitive complexity.} In fact, we can consider the order of the interactive concept {\small $S$} as the difficulty of a DNN modeling a concept.
% In contrast, people usually focus on the cognitive complexity, which is sometimes different from the learning difficulty of a DNN.
% For example, let us consider the recognition of a small ball concept consisting of a few pixels (\textit{low-order concept}) and a large ball concept consisting of massive pixels (\textit{high-order concept}) 
% in images.  
% People may consider these two balls have the same cognitive difficulty.
% However, from the perspective of a DNN, a large ball has more pixels, so that the DNN has to sophisticatedly examine whether all pixels within the large ball share the same color without exceptions. This is more difficult than examining a few pixels within a small ball. 
% Therefore, it may be more difficult for DNNs to recognize the large ball concept than the small ball concept.

\subsection{High-Order Concepts Are More Over-Fitted}
Although there is a common heuristic that complex concepts are usually more likely to be over-fitted, people still do not know the exact definition of concepts with an analytic connection to their generalization power. Because we also find the low generalization power of complex (high-order) interactive concepts, in this study, we make the first attempt to clarify the high inconsistency of complex (high-order) concepts, \emph{i.e.}, complex concepts are more sensitive to small noises in the data than simple concepts, which is responsible for the low generalization power of complex (high-order) concepts. Various experiments have verified our findings. This may shed new lights on how to evaluate the generalization power in terms of concepts.

\subsubsection{Illustrating Concepts of Different Orders.}
Before investigating the relationship between the complexity of concepts and the generalization power of a DNN, let us first visualize concepts extracted from a DNN. 
Specifically, we trained a seven-layer MLP (MLP-7-census) on the census dataset~\cite{census} and a seven-layer MLP (MLP-7-TV) on the TV news dataset~\cite{census}, respectively. Each layer of the MLPs contained 100 neurons.  
We also trained AlexNet~\cite{alexnet}, ResNet-20~\cite{resnet}, and VGG-11~\cite{vgg} on the MNIST dataset~\cite{mnist} (AlexNet-MNIST, ResNet-20-MNIST, VGG-11-MNIST) and the CIFAR-10 dataset~\cite{cifar10} (AlexNet-CIFAR-10, ResNet-20-CIFAR-10, VGG-11-CIFAR-10).

Although we only analyzed concepts in above DNNs in this very preliminary study, our findings could actually generalize to more diverse network architectures and datasets. It is because the proof for the sparse interactive concepts is agnostic to both network architectures and datasets.

\begin{figure*}[t]
   \centering
   \includegraphics[width=.9\linewidth]{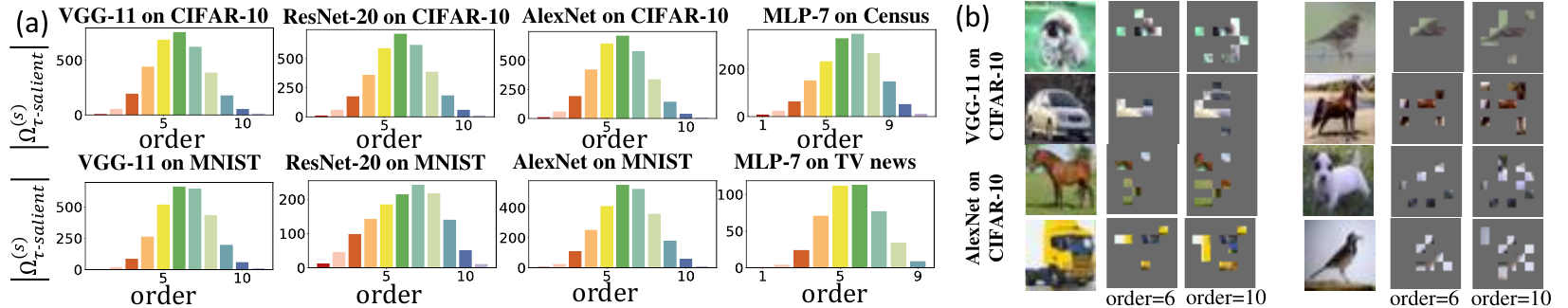}
   \caption{ (a) Histogram\footref{fn:bottleneck} of salient concepts of different orders, {\small $\vert \Omega_{\tau\text{-salient}} ^{(s)} \vert$}. (b) Visualization of salient concepts of different orders. Salient concepts are usually made up by image patches that contain discriminative parts of the object.}
   \label{fig:visualization}
\end{figure*}

\begin{figure*}[t]
    \centering
    \includegraphics[width=0.9\linewidth]{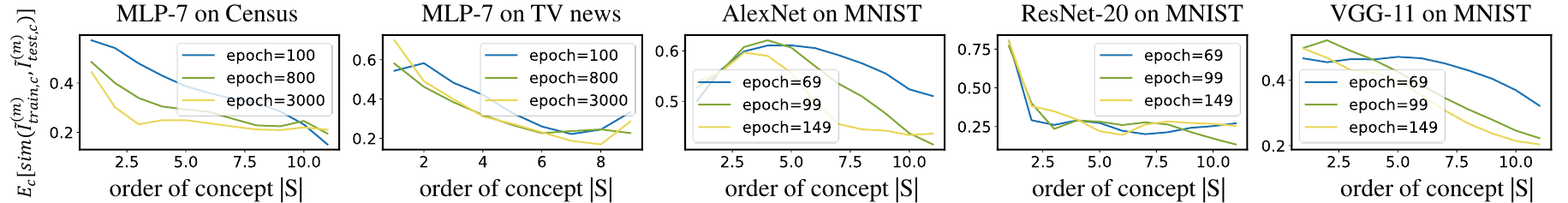}
    \caption{Average similarity between interactive concepts from training samples and those extracted from testing samples.}
  \label{fig:similarity_concept}
\end{figure*}

% \begin{figure*}[t]
%   \centering
%   \begin{minipage}{0.8\linewidth}
%       \centering
%       \includegraphics[width=\linewidth]{figures/similarity.pdf}
%   \end{minipage}%
%   \begin{minipage}{0.2\linewidth}
%       \caption{Average similarity between concepts from training samples and those extracted from testing samples.}
%   \label{fig:similarity_concept}
%   \end{minipage}
% \end{figure*}

Given a DNN and an input sample {\small $\boldsymbol{x} \in \mathbb{R}^n$}, we computed numerical effects {\small $I(S| \boldsymbol{x})$} of all {\small $2^n$} potential interactions, {\small $S\subseteq N$}.
We found a common phenomenon that interactions encoded by a DNN were usually very \emph{\textbf{sparse}}, which ubiquitously existed in different DNNs. 
As Figure~\ref{fig:sparsity} shows, numerical effects of 80\%-95\% interactive concepts were almost zero, {\small $\vert I(S \vert \boldsymbol{x})\vert \approx 0$}, and only a small number of interactive concepts had relatively significant effects {\small $\vert I(S \vert \boldsymbol{x})\vert$}. 
This finding was also consistent with the conclusion found by \citet{ren2023arrived}.
The sparsity of interactive concepts ensured the trustworthiness of such a concept definition.

Moreover, for each DNN, we compared the number of salient concepts of different orders.
Given a threshold {\small $\tau$}, we defined the set of salient concepts {\small $\Omega_{\text{salient}}$} as interactive concepts whose strengths were greater than {\small $\tau$}, \emph{i.e.}, {\small $\Omega_{\text{salient}} = \{S \subseteq N: \vert I(S \vert \boldsymbol{x}) \vert > \tau\}$}.  
Accordingly, {\small $\Omega_{\text{salient}}^{(s)} = \{S \subseteq N: \vert I(S \vert \boldsymbol{x}) \vert > \tau \text{ and } |S| = s\}$} represented all salient concepts of the $s$-th order.
Figure~\ref{fig:visualization} (a) shows the number of salient concepts of different orders, {\small $\vert \Omega_{\text{salient}} ^{(s)} \vert$}, where the threshold was set to be { \small $ \tau = 0.05 \cdot \max_S \vert I(S \vert \boldsymbol{x}) \vert$}. This figure illustrats that there were more middle-order salient concepts than low-order salient concepts and high-order salient concepts.\footnote{\label{fn:bottleneck}This conclusion does not conflict with, but actually supports, the representation bottleneck found by~\citet{deng2021bottleneck}, because that work used a different type of interaction. Please see Section 2 and Section 3 in the supplementary material for more discussion.}

Finally, in Figure~\ref{fig:visualization} (b), we visualized several salient concepts, covering middle-order and high-order concepts. We used the aforementioned experimental settings of AlexNet-CIFAR-10 and VGG-11-CIFAR-10, and we set {\small $\tau = 0.05 \cdot \max_{S} \vert I(S \vert \boldsymbol{x}) \vert $}. We found that salient concepts of different orders were usually made up by image patches that contained discriminative parts of the object.  
%要写哪些DNN，取怎样的threshold。可视化的concepts旁边也要标注"I(S|x)=??"。后面还有对着结果发表看法——比如各个阶的concepts的独有的特点——强度、稀疏度等等。

% \par % 开始新的段落
% \vspace{\baselineskip} % 插入空行
% \noindent\textbf{2.2.2 \quad Generalization to Testing Samples} 
% \vspace{2mm}\\
\subsubsection{Generalization to Testing Samples.}
Before the analytic explanation for the generalization power of a DNN, let us first experimentally verify that \textbf{compared to high-order interactive concepts, low-order interactive concepts are more likely to have the distribution in training samples being similar to the distribution in testing samples.}
To this end, previous studies used the gap of the loss~\cite{generalization_bounds,bousquet2020sharper,deng2021toward,haghifam2020sharpened, haghifam2021towards} or the smoothness of the loss landscape~\cite{flat_minima,loss_lanscape,foret2021sharpness,kwon2021asam} to investigate the generalization power of a DNN.

In comparison, the decomposition of interactive concepts provides us \textit{a more straightforward way} to define the generalization power of a DNN.
\emph{I.e.}, \textit{if an interactive concept is frequently extracted by the DNN from training samples, then it is also supposed to frequently appear in testing samples.
Otherwise, this interactive concept is not considered to be well generalized.}

In this way, we can define the generalization power of {\small $m$}-order interactive concepts \emph{w.r.t.} the category {\small $c$} as the similarity between the distribution of {\small $m$}-order interactive concepts in training samples of category {\small $c$} and that in testing samples of category {\small $c$}.
Let the vector {\small $I^{(m)}_{\text{train}, c}=[I^{(m)}_{\text{train}, c}(S_1),I^{(m)}_{\text{train}, c}(S_2),...,I^{(m)}_{\text{train}, c}(S_d)]^{\top} \in \mathbb{R}^d$} represent the distribution of {\small $m$}-order interactive concepts over training samples in the category {\small $c$}, which enumerates all {\small $d=\tbinom{n}{m}$} possible $m$-order interactive concepts. The $i$-th dimension {\small $I^{(m)}_{\text{train}, c}(S_i)= \mathbb{E}_{\boldsymbol{x} \in D_{\text{train},c}}[I(S_i \vert \boldsymbol{x})]$} represents the average effect of the interactive concept {\small $S_i$} over different training samples in the category {\small $c$}. Accordingly, the vector {\small $I^{(m)}_{\text{test}, c}$} denotes the distribution of $m$-order concepts over testing samples in the category {\small $c$}. Then, the similarity of the concept distribution between training samples and testing samples is given as the Jaccard similarity between {\small $\tilde{I}^{(m)}_{\text{train}, c}$} and {\small $\tilde{I}^{(m)}_{\text{test}, c}$},
\begin{equation}
\small
    \text{sim}(\tilde{I}^{(m)}_{\text{train}, c}, \tilde{I}^{(m)}_{\text{test}, c})
    = \frac{\Vert \min(\tilde{I}^{(m)}_{\text{train}, c}, \tilde{I}^{(m)}_{\text{test}, c}) \Vert_{1}}{\Vert \max(\tilde{I}^{(m)}_{\text{train}, c}, \tilde{I}^{(m)}_{\text{test}, c}) \Vert_{1}},
\label{eq:similarity}
\end{equation}
where we extend the {\small $d$}-dimensional vector {\small $I^{(m)}_{\text{train}, c}$} into a {\small $2d$}-dimensional vector {\small $\tilde{I}^{(m)}_{\text{train}, c} = [(I^{(m),+}_{\text{train}, c})^{\top}, (-I^{(m),-}_{\text{train}, c})^{\top}]^{\top} = [(\max(I^{(m)}_{\text{train}, c},0))^{\top}, (-\min(I^{(m)}_{\text{train}, c}, 0))^{\top}]^{\top} \in \mathbb{R}^{2d}$} with non-negative elements. Similarly, {\small $\tilde{I}^{(m)}_{\text{test}, c}$} is constructed on {\small $I^{(m)}_{\text{test}, c}$} to contain non-negative elements. 
Thus, a high similarity {\small $\text{sim}(\tilde{I}^{(m)}_{\text{train}, c}, \tilde{I}^{(m)}_{\text{test}, c})$} indicates that most {\small $m$}-order interactive concepts in the category {\small $c$} can be well generalized to testing samples in the category {\small $c$}.

\begin{figure}[t]
  \centering
  \includegraphics[width=1\linewidth]{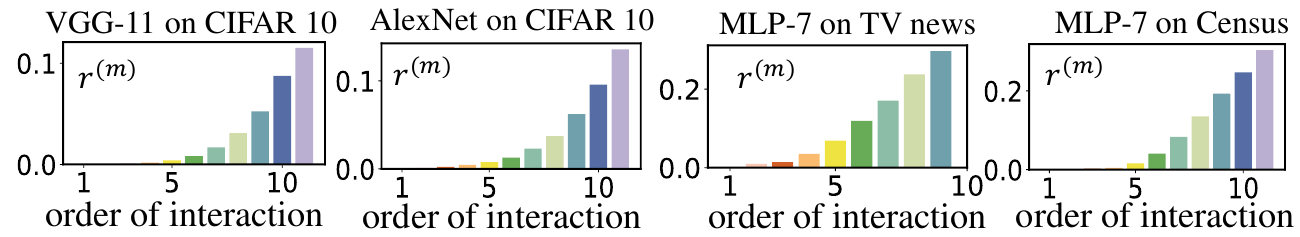}
  \caption{Comparison of the ratio $r^{(m)}$ of inconsistent concepts over different orders. High-order interactive concepts are usually more likely to make inconsistent effects on given noisy data, which verifies Theorem~\ref{theorem1}. }
  \label{fig:inconsistency}
\end{figure}

% \begin{figure}[t]
%   \centering
%   \begin{minipage}{0.65\linewidth}
%       \includegraphics[width=0.95\linewidth]{figures/inconsistency.pdf}
%   \end{minipage}%
%   \begin{minipage}{0.35\linewidth}
%       \caption{The ratio $r^{(m)}$ of inconsistent concepts over different orders. High-order concepts are more likely to make inconsistent effects on given noisy data, which verifies Theorem~\ref{theorem1}.}
%       \label{fig:inconsistency}
%   \end{minipage}
% \end{figure}

Furthermore, we conducted experiments to check whether high-order interactive concepts were more likely to be over-fitted.
Specifically, we trained a seven-layer MLP for the census dataset and the TV news dataset, respectively.
We also trained AlexNet, VGG-11 and ResNet-20 on the MNIST dataset.
Given each DNN, we computed interactive concepts of each order {\small $m$}. For each category {\small $c$}, we measured the above conceptual similarity for {\small $m$}-order interactive concepts, and Figure~\ref{fig:similarity_concept} reported the average similarity over different categories, \emph{i.e.}, {\small $\textit{similarity}=\mathbb{E}_c[\text{sim}(\tilde{I}^{(m)}_{\text{train}, c}, \tilde{I}^{(m)}_{\text{test}, c})]$}.
We found that low-order interactive concepts usually had more similar distributions between training data and testing data than high-order interactive concepts. This meant that compared to high-order concepts, the DNN was more likely to extract similar low-order concepts from the training data and testing data. In other words, low-order concepts in training data could be better generalized to testing data.

% \par % 开始新的段落
% \vspace{\baselineskip} % 插入空行
% \noindent\textbf{2.2.3 \quad Inconsistency of High-Order Concepts}
% \vspace{2mm}\\
\subsubsection{Inconsistency of High-Order Concepts.}
In this study, we try to use the inconsistency of high-order concepts to explain their low generalization power. Before the proof, this subsection first verifies that high-order concepts usually make inconsistent interactive effects on noisy data, \emph{i.e.}, the same high-order concept may push a sample towards a category, but pull another sample away from this category. Intuitively, such inconsistent effects over different samples usually makes an interactive concept perform like a noisy pattern, rather than a generalizable concept. 

Specifically, the inconsistency of a concept over different samples can be measured as follows.
Given a normal sample {\small $\boldsymbol{x}$}, we select a set of salient concepts of each {\small $m$}-th order, {\small $\Omega_{\boldsymbol{x}}^{(m)} = \{S \subseteq N: \vert S \vert = m \ \land \ \vert I(S\vert \boldsymbol{x}) \vert > \tau \}$}, the threshold is set to be {\small $\tau = 0.05 \cdot \text{max}_{S} \vert I(S\vert \boldsymbol{x})\vert$}. 
Then, by adding the adversarial perturbation generated by~\citet{adversarial_attacks}, we get an adversarial example {\small $\tilde{\boldsymbol{x}} = \boldsymbol{x} + \boldsymbol{\delta}$}. $I(S\vert \tilde{\boldsymbol{x}})$ denotes the interaction effect of the originally salient concepts $S \in \Omega_{\text{salient}}$ on the adversarial example $\tilde{x}$.
If {\small $I(S \vert \boldsymbol{x})$} and {\small $I(S \vert \tilde{\boldsymbol{x}})$} have the same sign, \emph{i.e.}, $I(S \vert \boldsymbol{x}) \cdot I(S \vert \boldsymbol{\Tilde{x}}) > 0$, we consider that the interactive concept {\small $S$} is consistent in adversarial attacking; otherwise not. 
Thus, we compute the ratio of inconsistent concepts to all the {\small $m$}-order salient concepts as {\small $r^{(m)} = \mathbb{E}_{\boldsymbol{x} \in D} \frac{\vert \{S \subseteq N: S \in \Omega_{\boldsymbol{x}}^{(m)} \ \land \ I(S \vert \boldsymbol{x}) \cdot I(S \vert \boldsymbol{\Tilde{x}}) < 0 \} \vert}{\vert \Omega_{\boldsymbol{x}}^{(m)} \vert}$}.
% , and examine the relationship between $r^{(m)}$ and the order of the concept.

To compare the inconsistency over different orders, we follow experimental settings in the Section 2.2.1 to train AlexNet and VGG-11 on the CIFAR-10 dataset, and to train seven-layer MLPs on the census dataset and the TV news dataset. Figure~\ref{fig:inconsistency} shows that the ratio {\small $r^{(m)}$} of inconsistent salient concepts increases along with the order {\small $m$}.

% \par % 开始新的段落
% \vspace{\baselineskip} % 插入空行
% \noindent\textbf{2.2.4 \quad Analytic Inconsistency of Concepts}
% \vspace{2mm}\\

\subsubsection{Analytic Inconsistency of Concepts.}
All above experimental findings on the generalization power of concepts are related to the phenomenon of the inconsistency of high-order concepts, \emph{i.e.}, high-order concepts are more sensitive to small noises in the input sample than low-order concepts.
Therefore, we aim to prove that \textbf{the interaction effect’s variance of the concept increases with the concept's order exponentially under a simple setting.}
Let us add a Gaussian perturbation {\small $\boldsymbol{\epsilon} \sim \mathcal{N}(\boldsymbol{0}, \delta^2\boldsymbol{I})$} to the input sample {\small $\boldsymbol{x}$} and obtain {\small $\boldsymbol{x}^\prime = \boldsymbol{x} + \boldsymbol{\epsilon}$}. 
The added perturbation represents noises/variations that inevitably exist in the data.
We admit that there are other types of noises, such as texture variations and the shape deformation in object classification. In this study, we just use the Gaussian perturbation to represent the noises/variations in the data. Our conclusion may still provide conceptual insights into real-world applications.
%Actually, noises/variations inevitably exist in the data, such as chaotic textures on an object or the scale/position change for an object during object detection, which are usually difficult to model. Therefore, we use the Gaussian perturbation to represent the noises/variations in the data for simplicity. 
% Then, Theorem~\ref{theorem1} reformulates the interactive effect {\small $I(S\vert \boldsymbol{x}^\prime)$} using the Taylor series expansion.

\begin{lemma}\label{lemma:Taylor expansion}
Given a neural network {\small $v$} and an arbitrary perturbed input sample {\small $\boldsymbol{x}^\prime = \boldsymbol{x} + \boldsymbol{\epsilon}$}, the neural network output {\small $v(\boldsymbol{x}^\prime)$} can be rewritten by following the Taylor series expansion at the baseline point {\small $\boldsymbol{b} = [b_1, \dots, b_n]^{\top}$},
\begin{equation}
\begin{aligned}
\small
v(\boldsymbol{x}^\prime) &= v(\boldsymbol{b}) + \sum_{k=1}^{\infty}\sum_{\boldsymbol{\kappa} \in O_k} C(\boldsymbol{\kappa}) \cdot \nabla_v(\boldsymbol{\kappa}) \cdot  \pi(\boldsymbol{\kappa}|\boldsymbol{x}^\prime),\\
\end{aligned}
\end{equation}
including the coefficient {\small $C(\boldsymbol{\kappa}) = \frac{1}{(\kappa_1 + \dots + \kappa_n)!} \tbinom{\kappa_1 + \dots + \kappa_n}{\kappa_1,\dots,\kappa_n}$ $\in \mathbb{R}$}, the partial derivative {\small $\nabla_v(\boldsymbol{\kappa}) = \frac{\partial^{\kappa_1+\dots+\kappa_n}v(\boldsymbol{b})}{\partial^{\kappa_1}x_1 \dots \partial^{\kappa_n}x_n} \in \mathbb{R}$}, and the expansion term
{\small $\pi(\boldsymbol{\kappa}|\boldsymbol{x}^\prime) = \prod\nolimits_{i = 1}^n (x_i^\prime - b_i)^{\kappa_i}$}. 
Here, {\small $\boldsymbol{\kappa} = [\kappa_1, \dots, \kappa_n] \in \mathbb{N}^n$} denotes the non-negative integer degree vector of each Taylor expansion term.
Correspondingly, {\small $O_k = \{\boldsymbol{\kappa} \in \mathbb{N}^n | \kappa_1 + \dots + \kappa_n = k\}$} represents the set of all expansion terms of the {\small $k$}-th order. 
\end{lemma}

As a prerequisite, Lemma \ref{lemma:Taylor expansion} gives the Taylor series expansion when the network output {\small $v(\boldsymbol{x}^\prime)$}  is expanded at the baseline point {\small $\boldsymbol{b}  = [b_1, \dots, b_n]^{\top}$}. 
We use the baseline value {\small $b_i$} to represent the masking state of the input variable {\small $x_i$}.
Normally, we can set the input variable {\small $x_i$} as the average value of $x_i$ over different samples to remove the information~\cite{no_signal}, \textit{i.e.}, {\small $b_i = \mu_i = \mathbb{E}_{\boldsymbol{x}}[x_i]$}. However, pushing the input variable {\small $x_i$} a big distance {\small $\alpha\in \mathbb{R}$} towards {\small $\mu_i$} is usually enough to remove the information in real applications. 
Thus, we temporarily set {\small $b_i = x_i+ \alpha$}, if {\small $x_i < \mu_i$}; and set {\small $b_i= x_i - \alpha$}, if {\small $x_i > \mu_i$}, to simplify the proof.

\begin{theorem}
Given a neural network {\small $v$} and an arbitrary perturbed input sample {\small $\boldsymbol{x}^\prime = \boldsymbol{x}+\boldsymbol{\epsilon}$} by adding a Gaussian perturbation {\small $\boldsymbol{\epsilon} \sim \mathcal{N}(\boldsymbol{0}, \delta^2 \boldsymbol{I})$}, the interactive effect {\small $I(S\vert \boldsymbol{x}^\prime)$} is defined by setting {\small $(\boldsymbol{x}^\prime_T)_i = x^\prime_i$} if {\small $i \in T$} and setting {\small $(\boldsymbol{x}^\prime_T)_i = b_i$ if $i \not\in T$}. Then, we obtain 
\begin{equation}
    \begin{aligned}
    \small
     I(S \vert \boldsymbol{x}^\prime) &= \sum\nolimits_{\boldsymbol{\kappa} \in Q_S} C(\boldsymbol{\kappa}) \cdot \nabla_v(\boldsymbol{\kappa}) \cdot  \pi(\boldsymbol{\kappa}|\boldsymbol{x}^\prime) \\
     &= \sum\nolimits_{\boldsymbol{\kappa} \in Q_S} Z(\boldsymbol{\kappa}) \cdot \hat \pi(\boldsymbol{\kappa}|\boldsymbol{x}^\prime),
    %&=  \sum_{\boldsymbol{\kappa} \in Q_S}  \left(\frac{1}{Z_{\boldsymbol{\kappa}}} C(\boldsymbol{\kappa})\nabla_v(\boldsymbol{\kappa}) \right) \cdot \underbrace{Z_{\boldsymbol{\kappa}} \ \pi(\boldsymbol{\kappa}|\boldsymbol{x}^\prime)}_{\text{\rm term } \hat \pi(\boldsymbol{\kappa}|\boldsymbol{x}^\prime)}
    \label{eq:continuous_I}
    \end{aligned}
\end{equation}
where {\small $\hat \pi(\boldsymbol{\kappa}|\boldsymbol{x}^\prime) =  \prod_{i=1}^n(\frac{\text{sign}(x_i-b_i)}{\alpha})^{\kappa_i} \cdot \pi(\boldsymbol{\kappa}|\boldsymbol{x}^\prime)$} is a standard AND interaction of the degree vector {\small $\boldsymbol{\kappa}$}, and it is normalized to satisfy {\small $\forall  \boldsymbol{\kappa} \in Q_S, \mathbb{E}_{\boldsymbol{\epsilon}}[\hat \pi(\boldsymbol{\kappa}|\boldsymbol{x}^\prime = \boldsymbol{x} + \boldsymbol{\epsilon})] = 1 + O(\delta^2)$}.
In addition,  
{\small $Z(\boldsymbol{\kappa}) =\prod_{i=1}^n(\frac{\alpha}{\text{sign}(x_i-b_i)})^{\kappa_i}  \cdot C(\boldsymbol{\kappa}) \cdot \nabla_v(\boldsymbol{\kappa})$} denotes the scalar coefficient for {\small $\hat \pi(\boldsymbol{\kappa}|\boldsymbol{x}^\prime)$.
$Q_S = \{\boldsymbol{\kappa} \in \mathbb{N}^n \vert \forall i \in S, \kappa_i \in \mathbb{N}^{+}; \forall i \notin S, \kappa_i=0 \}$} denotes the set of degree vectors corresponding to all Taylor expansion terms involving only variables in {\small $S$}.  
Furthermore, the second-order moment of the standard AND interaction
{\small $\hat \pi(\boldsymbol{\kappa}|\boldsymbol{x} + \boldsymbol{\epsilon})$} w.r.t. the Gaussian perturbations {\small $\boldsymbol{\epsilon}$} is derived as follows.
\begin{equation}
\begin{aligned}
\small
\forall \boldsymbol{\kappa} \in Q_S, \mathbb{E}_{\boldsymbol{\epsilon}}[\hat \pi^2(\boldsymbol{\kappa}|\boldsymbol{x} + \boldsymbol{\epsilon})] = \prod_{i \in S} [1+ \sum_{m = 1}^{\kappa_i} c_m U_{2m}(\epsilon_i)],
   \label{eq:variance_J_}
   \end{aligned}
\end{equation}
where {\small $U_{2m}(\epsilon_i) = \mathbb{E} [\epsilon_i^{2m}]>0$, and $c_m = \tbinom{2 \kappa_i}{2m} \frac{1}{\alpha^{2m}}> 0$}. 
\label{theorem1}
\end{theorem}

\begin{figure}[t]
  \centering
  \includegraphics[width=0.75\linewidth]{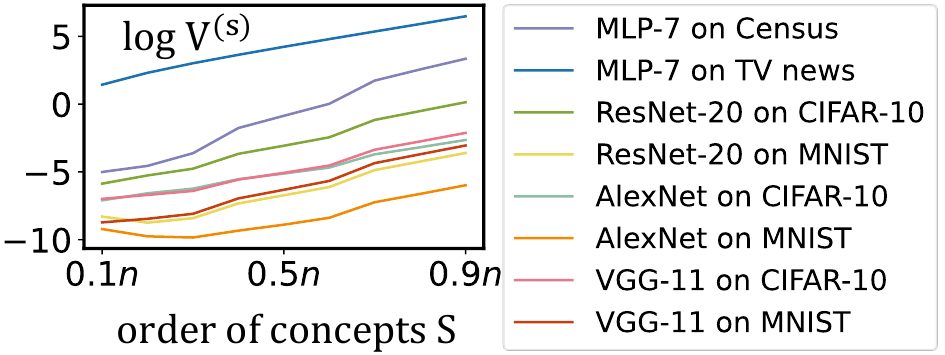}
  \caption{ Logarithm of the variance. The interactive effect's variance increased along with the order exponentially. It verifies the inconsistency of high-order concepts in Theorem~\ref{theorem1}.}
  \label{fig:noise_var}
\end{figure}

% \begin{figure}[t]
%   \centering
%   \begin{minipage}{0.7\linewidth}
%       \includegraphics[width=\linewidth]{figures/noise_var.pdf}
%   \end{minipage}%
%   \begin{minipage}{0.3\linewidth}
%       \caption{Logarithm of the variance, which verifies the inconsistency of high-order concepts.}
%   \label{fig:noise_var}
%   \end{minipage}
% \end{figure}

\textbf{Theorem~\ref{theorem1} reformulates the interactive effect $I(S \vert \boldsymbol{x}^\prime)$ into elementary standard interactions {\small $\hat \pi(\boldsymbol{\kappa}|\boldsymbol{x}^\prime)$}.} Specifically, Equation (\ref{eq:variance_J_}) tells us that for a specific standard interaction {\small $\hat \pi(\boldsymbol{\kappa}|\boldsymbol{x}^\prime)$} subject to {\small $\boldsymbol{\kappa} \in Q_S$},  its second-order moment increases along with the order {\small $\vert S \vert$} in a roughly exponential manner, but its mean value {\small $\mathbb{E}_{\boldsymbol{\epsilon}}[\hat \pi(\boldsymbol{\kappa}|\boldsymbol{x} + \boldsymbol{\epsilon})] \approx 1$} is independent with the order. 
Therefore, we can roughly consider that its variance {\small $\textit{Var}_{\boldsymbol{\epsilon}} [\hat \pi(\boldsymbol{\kappa}|\boldsymbol{x} + \boldsymbol{\epsilon})] =  \mathbb{E}_{\boldsymbol{\epsilon}}[\hat \pi^2(\boldsymbol{\kappa}|\boldsymbol{x} + \boldsymbol{\epsilon})] - \mathbb{E}^2_{\boldsymbol{\epsilon}}[\hat \pi(\boldsymbol{\kappa}|\boldsymbol{x} + \boldsymbol{\epsilon})]$} also increases along with the order {\small $\vert S \vert$} exponentially. 

Moreover, according to Equation~\eqref{eq:continuous_I}, the interactive effect {\small $I(S \vert \boldsymbol{x}^\prime)$} of the concept $S$ is the weighted sum of all elementary terms {\small $\hat \pi(\boldsymbol{\kappa}|\boldsymbol{x} + \boldsymbol{\epsilon})$} satisfying {\small $\boldsymbol{\kappa} \in Q_S$}, and different terms of {\small $\hat \pi(\boldsymbol{\kappa}|\boldsymbol{x} + \boldsymbol{\epsilon})$} \emph{w.r.t.} the same set $S$ are roughly positively correlated to each other. Thus, in the simple setting of adding Gaussian perturbations to the input, we can consider that the variance of {\small $I(S \vert \boldsymbol{x}^\prime)$} has approximately an exponent relation with the order {\small $\vert S \vert$} of the concept {\small $S$}. \textbf{Therefore, Theorem~\ref{theorem1} shows that high-order concepts usually make more inconsistent effects than low-order concepts.} Although there are other types of noises in the real data, our theory may still provide conceptual insights into real-world applications.

\textit{Experimental verification of Theorem~\ref{theorem1}.} We conducted experiments to verify the exponential relation of the interactive effect's variance with the order of the concept, which is predicted by Theorem~\ref{theorem1}.
To this end, given a well-trained DNN and an input sample {\small $\boldsymbol{x}$}, we added a Gaussian perturbation {\small $\boldsymbol{\epsilon} \sim \mathcal{N}(\boldsymbol{0}, \delta^2 \boldsymbol{I})$} to the input sample. 
Then, we used {\small $V^{(s)} = \mathbb{E}_{\boldsymbol{x}}[\mathbb{E}_{|S| = s} [\textit{Var}_{\boldsymbol{\epsilon}}[I(S \vert \boldsymbol{x} + \boldsymbol{\epsilon})]]]$} to measure the average variance of {\small $s$}-th order concepts \emph{w.r.t.} the Gaussian perturbation {\small $\boldsymbol{\epsilon}$}. 
In experiments, we used DNNs introduced in Section 2.2.1 for testing. Figure~\ref{fig:noise_var} shows that the interactive effect's variance {\small $V^{(s)}$} increased along with the order {\small $s$} in a roughly exponential manner.
% Considering that the mean value {\small $\mathbb{E}_{\boldsymbol{\epsilon}}[I(S \vert \boldsymbol{x} + \boldsymbol{\epsilon})]$} of {\small $I(S \vert \boldsymbol{x} + \boldsymbol{\epsilon})$} kept stable over different orders in Theorem~\ref{theorem1}, 
The inconsistency of high-order concepts in Theorem~\ref{theorem1} is verified.

\subsection{An Over-Fitted DNN Usually Encodes Strong High-Order Interactive Concepts}
\label{sec: over-fitted DNN}
In this subsection, we further analyze and explain the generalization power of the entire DNN based on the generalization power of the encoded interactive concepts.

To this end, we need to construct DNNs with different generalization power for investigation.  
In fact, the generalization power of a DNN is usually affected by various factors, such as the network architecture and training data.
In this study, we consider a typical case, \emph{i.e.}, random labels in training data usually push the DNN to be over-fitted to non-generalizable features for classification~\cite{noisy}.
Therefore, in experiments, we trained DNNs by applying different ratios of noise data.
Specifically, we trained a DNN on training samples with a {\small $\rho$ $(0 \leq \rho \leq 1)$} ratio of incorrect labels, which was termed a \textit{DNN with $\rho$ noise}. 
We considered that a DNN trained with more incorrect labels (a high $\rho$ value) was more over-fitted. 
We trained \textit{AlexNet, ResNet-20 and VGG-11 with $\rho=0,0.05,0.1,0.2,0.3$ noise} on the MNIST dataset and the CIFAR-10 dataset.

\begin{figure}[t]
\centering
\includegraphics[width=1\linewidth]{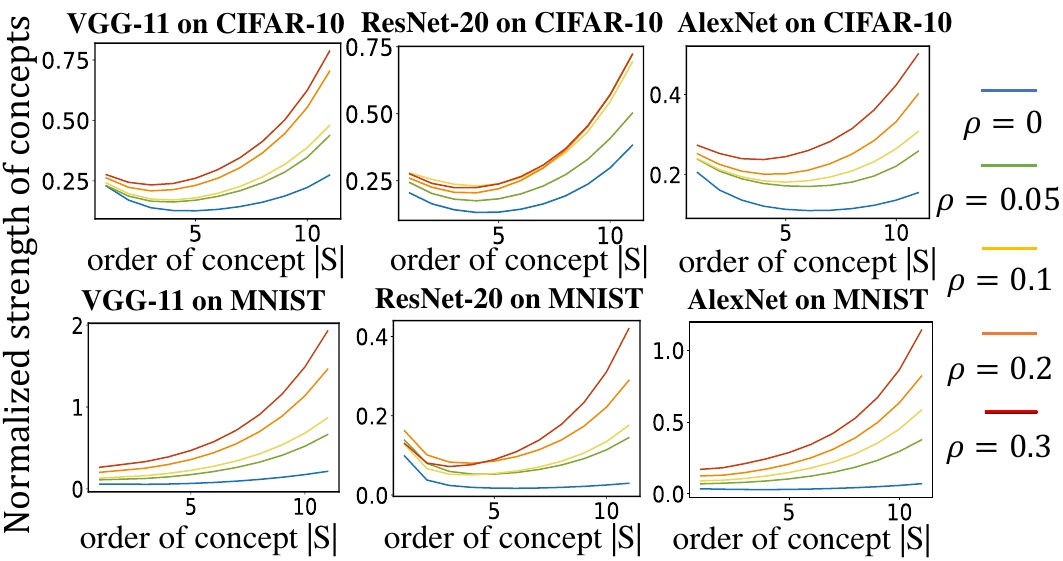}
\caption{Interaction strength of DNNs trained with different noise levels \textit{\small$\rho$}. We normalized the strength of interactive concepts {\small $\mathbb{E}_{\boldsymbol{x} \in D}\mathbb{E}_{S:\vert S \vert=m} [\frac{\vert I^{(m)}(S \vert \boldsymbol{x}) \vert}{\mathbb{E}_{\boldsymbol{x} \in D}[\vert v(N \vert \boldsymbol{x}) - v(\varnothing \vert \boldsymbol{x})]}]$}.}
%\caption{Normalized interaction strength of interactive concepts of different orders {\small $\mathbb{E}_{\boldsymbol{x} \in D}\mathbb{E}_{S:\vert S \vert=m} [\frac{\vert I^{(m)}(S \vert \boldsymbol{x}) \vert}{\mathbb{E}_{\boldsymbol{x} \in D}[\vert v(N \vert \boldsymbol{x}) - v(\varnothing \vert \boldsymbol{x})]}]$}.  We compare interaction strength between DNNs trained with different noise levels \textit{\small$\rho$}. }
% \textit{$\rho=0,0.05,0.1,0.2,0.3$}.}
\label{fig:interaction_strength}
\end{figure}

\textbf{Claim 1: More label noise usually makes DNNs to encode stronger high-order interactive concepts}.
\quad To verify this claim, for each DNN trained with a $\rho$ ratio of incorrect labels, we computed the average interaction strength of $m$-order interactive concepts over different training samples, \emph{i.e.}, {\small $\mathbb{E}_{\boldsymbol{x} \in D}\mathbb{E}_{S:\vert S \vert=m} [\frac{\vert I^{(m)}(S \vert \boldsymbol{x}) \vert}{\mathbb{E}_{\boldsymbol{x} \in D}[\vert v(N \vert \boldsymbol{x}) - v(\varnothing \vert \boldsymbol{x})]}]$.}
%To verify the above claim, we trained DNNs with different generalization power, which suffered from the over-fitting problem at different levels. In this way, we examined whether the DNN {\color{blue}trained with more label noise} will encode stronger high-order interactive concepts than the DNN {\color{blue}trained with less label noise}.
%Specifically, we assigned a ratio $\rho (0 \leq \rho \leq 1)$ of training samples with incorrect labels to train a DNN, which was termed a \textit{DNN with $\rho$ noise}. We considered that a DNN trained with more incorrect labels (a high $\rho$ value) was more over-fitted. In this way, we trained DNNs with different generalization powers by applying different ratios of incorrect labels.We trained \textit{AlexNet, ResNet-20 and VGG-11 with $\rho=0,0.05,0.1,0.2,0.3$ noise} on the MNIST dataset and the CIFAR-10 dataset.
% 实验设置：我们训练了哪些DNNs，哪些数据库。三个DNN，三个数据集。MNIST，CIFAR，TinyImageNet。
%Then, for each DNN, we computed the average interaction strength of concepts of each specific order $m$ over different training samples, \emph{i.e.}, $\mathbb{E}_{\boldsymbol{x} \in D}\mathbb{E}_{S:\vert S \vert=m} [\frac{\vert I^{(m)}(S \vert \boldsymbol{x}) \vert}{\mathbb{E}_{\boldsymbol{x} \in D}[\vert v(N \vert \boldsymbol{x}) - v(\varnothing \vert \boldsymbol{x})]}]$.
Figure~\ref{fig:interaction_strength} shows that the DNN trained with more incorrect labels (a high {\small $\rho$} value) usually encoded more significant high-order concepts than the DNN trained with fewer incorrect labels. 
In other words, DNNs trained with more label noise (with poorer generalization power) usually encoded stronger high-order concepts.

\begin{figure}[t]
\centering
\includegraphics[width=1\linewidth]{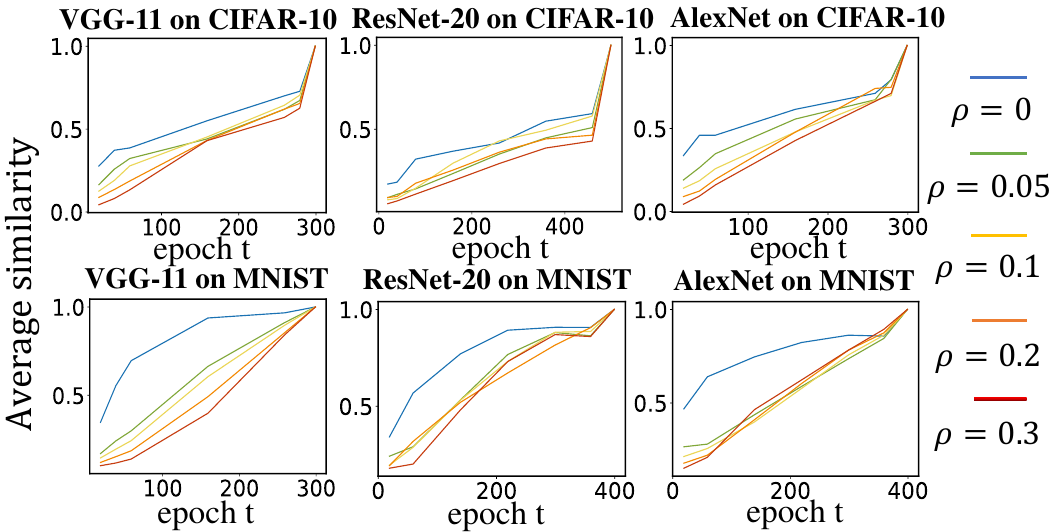}
\caption{Average similarity {\small $\text{Sim}^{(m=1,t)}$} between low-order interactive concepts encoded after the {\small $t$}-th epoch and those encoded after all epochs. We compare {\small $\text{Sim}^{(m=1,t)}$} between DNNs trained with different noise levels \textit{$\rho$}. More results can be found in supplemental materials.}
% The results for 2-order concepts and 3-order concepts are shown in Appendix.}
\label{fig:iou}
\end{figure}

\textbf{Claim 2: Less label noise usually makes DNNs to learn low-order interactive concepts more quickly.}
\quad To verify this claim, for each DNN trained with a {\small $\rho$} ratio of incorrect labels, we examined the learning progress of interactive concepts of a specific order {\small $m$}.
We used the metric {\small $\text{Sim}^{(m,t)} = \mathbb{E}_{\boldsymbol{x} \in D}[\text{sim}(\tilde{I}^{(m)}_t(\boldsymbol{x}), \tilde{I}^{*(m)}(\boldsymbol{x}))]$} to measure the learning progress of {\small $m$}-order interactive concepts at the {\small $t$}-th epoch, which was defined as the average Jaccard similarity between all {\small $m$}-order interactive concepts {\small $\tilde{I}^{*(m)}(\boldsymbol{x})$} encoded by the finally trained DNN and all {\small $m$}-order interactive concepts {\small $\tilde{I}^{(m)}_t(\boldsymbol{x})$} encoded by the DNNs {\small $v^{(t)}$} trained after {\small $t$} epochs.
Here, {\small $\text{sim}(\cdot)$} and the two vectors {\small $\tilde{I}^{*(m)}(\boldsymbol{x})$}, {\small $\tilde{I}^{(m)}_t(\boldsymbol{x})$} were defined in Equation~\eqref{eq:similarity}.
In this way, if a DNN obtained a high similarity {\small $\text{Sim}^{(m,t)}$} (\textit{i.e.}, achieved a high learning progress) in early epochs, then we considered that this DNN learned {\small $m$}-order interactive concepts quickly. 

Then, we conducted experiments to compare the learning speeds of interactive concepts between aforementioned DNNs with different ratios of label noise.
Figure~\ref{fig:iou} shows that a DNN trained with less label noise usually exhibited a higher {\small $\text{Sim}^{(m,t)}$} for low-order interactive concepts.
It meant that DNNs trained with less label noise usually learned low-order interactive concepts more quickly.

\subsection{Detouring Dynamics of High-Order Concepts}
In this section, we analyze the learning dynamics of concepts with a simple experimental setting, \emph{i.e.}, using a DNN to fit a boolean polynomial. We find that a high-order concept is not directly learned, but is likely to be mistakenly encoded as a mixture of low-order concepts in early epochs. In spite of the simplicity of experiments, this finding may still provide conceptual insights into the reason why high-order concepts are more likely to be over-fitted.

\begin{figure}[t]
    \centering
    \includegraphics[width=0.95\linewidth]{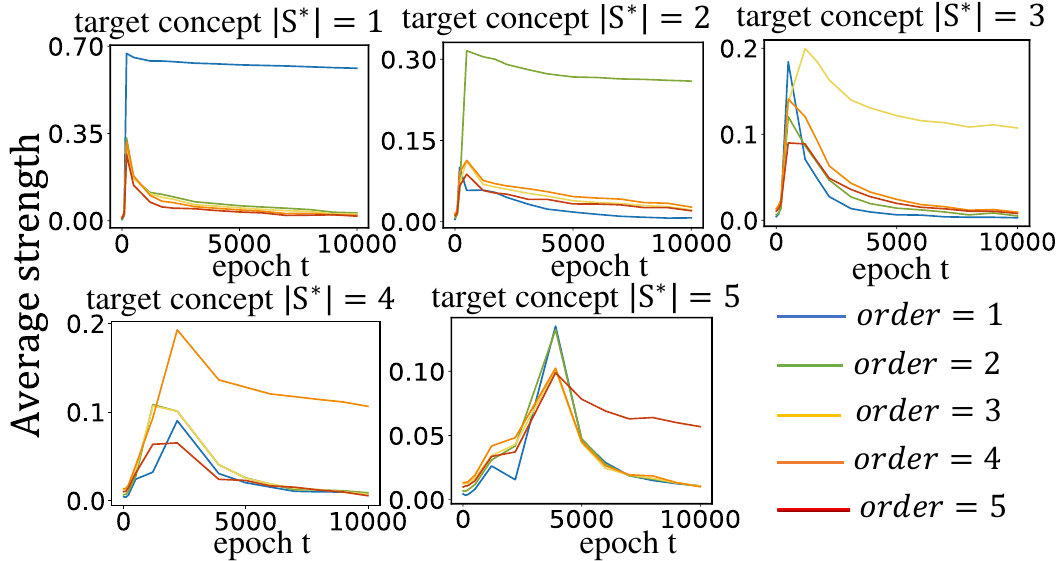}
    \caption{Average strength {\small $\mathbb{E}_{\boldsymbol{x} \in X} [\sum\nolimits_{S: \vert S \vert =m}\vert I^{(m)}_{t}(S \vert \boldsymbol{x}) \vert]$} of interactive concepts of different orders.}
   \label{fig:learning_dynamic}
\end{figure}

We trained a DNN {\small $v$} to fit a random concept {\small $S^{*} \subseteq N$} of the {\small $m$}-th order, {\small $m=\vert S^{*}\vert$}. Given an arbitrary input sample {\small $\boldsymbol{x}$}, if the input variable {\small $x_i$} was set to the original value, then we set {\small $A_i=1$}; if the input variable was masked, then we set {\small $A_i=0$}. Thus, the {\small $m$}-order target concept was formulated as {\small $u_{S^{*}}(\boldsymbol{x}) = \prod_{i\in S^{*}}A_i$}. We trained the DNN to fit the function of the concept {\small $u_{S^{*}}(\boldsymbol{x})$} based on the {\small $Loss=E_{\boldsymbol{x} \in X} [\Vert v(\boldsymbol{x})-u_{S^{*}}(\boldsymbol{x}) \Vert^2]$}, \emph{i.e.}, to fit a boolean polynomial. We trained a five-layer MLP on the dataset {\small $X=\{ 0,1 \}^n$}, which contained samples corresponding to all masking states {\small $\boldsymbol{x}=[x_1,x_2,\ldots,x_n]^{\top} \in X, \forall i, x_i \in \{0,1\}$}. Here, {\small $n=10$}.

If a DNN had been well trained after {\small $t$} epochs, then it was supposed to only extract a single concept with non-zero effect {\small $I_t^{(m)}(S \vert \boldsymbol{x})$}. 
Therefore, we examined whether the DNN encoded concepts of different orders or a single concept.

To this end, we used {\small $\mathbb{E}_{\boldsymbol{x} \in X} [\sum\nolimits_{S: \vert S \vert =m}\vert I^{(m)}_{t}(S \vert \boldsymbol{x}) \vert]$} to denote the average strength of $m$-order interactions after the $t$-th epoch. We tracked the change of the average interaction strength over different epochs.
% 然后写做了哪些DNN，用了哪些数据集。
Figure~\ref{fig:learning_dynamic} shows that when a DNN was trained to fit a low-order concept, it usually learned such a concept directly.
In comparison, when a DNN was trained to fit a high-order concept, the learning dynamics is detouring. Specifically, the DNN usually first learned low-order concepts. 
Then, the DNN shifted its attention to concepts of higher orders, and later gradually removed mistakenly learned low-order concepts.

In this way, the above detouring dynamics of high-order concepts showed that high-order concepts were more difficult to be learned.
A high-order concept was likely to be mistakenly encoded as a mixture of low-order concepts. 
Therefore, high-order concepts were less likely to be generalized to testing data than low-order concepts.

\section{Conclusion, discussions, and future challenges}
\label{sec:conclusion}
%-------------------------------------------------------------------------
In this paper, we provide a conceptual understanding of the reason why low-order concepts in training data can usually better generalize to testing data than high-order concepts. Specifically, we prove that the average inconsistency of concepts usually increases exponentially along with the order of concepts. We find that DNNs with poorer generalization power usually encode more high-order concepts, and DNNs with stronger generalization power usually encode low-order concepts more quickly. Moreover, we find that low-order concepts are usually learned directly, but high-order concepts are more likely to be mistakenly encoded as a mixture of various incorrect low-order concepts. These all explain the low generalization power of high-order interactive concepts. Section 8 in supplemental materials will introduce future practical values of this study.

In fact, towards the problem of faithfully explaining decision-making logic of a DNN into symbolic inference patterns, a theory system of interaction-based explanation has been constructed with over 20 papers, as surveyed by \cite{ren2023arrived}. Despite these achievements, it is still hard to say the interaction is \textit{``the first-principal explanation''} of the DNN. There are still several unsolved big scientific problems in this field, \emph{e.g.,} how to use interactions to faithfully summarize the complex learning dynamics of a DNN, and how to learn from interactions extracted from a DNN.

\section*{Acknowledgments}
This work is partially supported by the National Key R\&D Program of China (2021ZD0111602), the National Nature Science Foundation of China (62276165), Shanghai Natural Science Foundation (21JC1403800, 21ZR1434600) and the National Nature Science Foundation of China (62206170). This work is also partially supported by Huawei Technologies Inc.

\bibliography{aaai24}

\begin{thebibliography}{31}
\providecommand{\natexlab}[1]{#1}

\bibitem[{Ancona, Oztireli, and Gross(2019)}]{no_signal}
Ancona, M.; Oztireli, C.; and Gross, M. 2019.
\newblock Explaining deep neural networks with a polynomial time algorithm for
  shapley value approximation.
\newblock In \emph{International Conference on Machine Learning}, 272--281.
  PMLR.

\bibitem[{Asuncion and Newman(2007)}]{census}
Asuncion, A.; and Newman, D. 2007.
\newblock UCI machine learning repository.

\bibitem[{Bae et~al.(2022)Bae, Shin, Na, Jang, Song, and Moon}]{noisy}
Bae, H.; Shin, S.; Na, B.; Jang, J.; Song, K.; and Moon, I.-C. 2022.
\newblock From noisy prediction to true label: Noisy prediction calibration via
  generative model.
\newblock In \emph{International Conference on Machine Learning}, 1277--1297.
  PMLR.

\bibitem[{Blumer et~al.(1987)Blumer, Ehrenfeucht, Haussler, and
  Warmuth}]{occam}
Blumer, A.; Ehrenfeucht, A.; Haussler, D.; and Warmuth, M.~K. 1987.
\newblock Occam's razor.
\newblock \emph{Information processing letters}, 24(6): 377--380.

\bibitem[{Bousquet, Klochkov, and Zhivotovskiy(2020)}]{bousquet2020sharper}
Bousquet, O.; Klochkov, Y.; and Zhivotovskiy, N. 2020.
\newblock Sharper bounds for uniformly stable algorithms.
\newblock In \emph{Conference on Learning Theory}, 610--626. PMLR.

\bibitem[{Cheng et~al.(2021{\natexlab{a}})Cheng, Chu, Zheng, Ren, and
  Zhang}]{chengxv}
Cheng, X.; Chu, C.; Zheng, Y.; Ren, J.; and Zhang, Q. 2021{\natexlab{a}}.
\newblock A game-theoretic taxonomy of visual concepts in dnns.
\newblock \emph{arXiv preprint arXiv:2106.10938}.

\bibitem[{Cheng et~al.(2021{\natexlab{b}})Cheng, Wang, Xue, Liang, and
  Zhang}]{cheng2021proto}
Cheng, X.; Wang, X.; Xue, H.; Liang, Z.; and Zhang, Q. 2021{\natexlab{b}}.
\newblock A hypothesis for the aesthetic appreciation in neural networks.
\newblock \emph{arXiv preprint arXiv:2108.02646}.

\bibitem[{Dabkowski and Gal(2017)}]{baseline_mean}
Dabkowski, P.; and Gal, Y. 2017.
\newblock Real time image saliency for black box classifiers.
\newblock \emph{Advances in neural information processing systems}, 30.

\bibitem[{Deng et~al.(2021)Deng, Ren, Zhang, and Zhang}]{deng2021bottleneck}
Deng, H.; Ren, Q.; Zhang, H.; and Zhang, Q. 2021.
\newblock DISCOVERING AND EXPLAINING THE REPRESENTATION BOTTLENECK OF DNNS.
\newblock In \emph{International Conference on Learning Representations}.

\bibitem[{Deng, He, and Su(2021)}]{deng2021toward}
Deng, Z.; He, H.; and Su, W. 2021.
\newblock Toward better generalization bounds with locally elastic stability.
\newblock In \emph{International Conference on Machine Learning}, 2590--2600.
  PMLR.

\bibitem[{Dziugaite and Roy(2017)}]{pac_bound}
Dziugaite, G.~K.; and Roy, D.~M. 2017.
\newblock Computing nonvacuous generalization bounds for deep (stochastic)
  neural networks with many more parameters than training data.
\newblock \emph{arXiv preprint arXiv:1703.11008}.

\bibitem[{Foret et~al.(2021)Foret, Kleiner, Mobahi, and
  Neyshabur}]{foret2021sharpness}
Foret, P.; Kleiner, A.; Mobahi, H.; and Neyshabur, B. 2021.
\newblock Sharpness-aware minimization for efficiently improving
  generalization.
\newblock In \emph{International Conference on Learning Representations}.

\bibitem[{Fort et~al.(2019)Fort, Nowak, Jastrzebski, and Narayanan}]{stiffness}
Fort, S.; Nowak, P.~K.; Jastrzebski, S.; and Narayanan, S. 2019.
\newblock Stiffness: A new perspective on generalization in neural networks.
\newblock \emph{arXiv preprint arXiv:1901.09491}.

\bibitem[{Haghifam et~al.(2021)Haghifam, Dziugaite, Moran, and
  Roy}]{haghifam2021towards}
Haghifam, M.; Dziugaite, G.~K.; Moran, S.; and Roy, D. 2021.
\newblock Towards a unified information-theoretic framework for generalization.
\newblock \emph{Advances in Neural Information Processing Systems}, 34:
  26370--26381.

\bibitem[{Haghifam et~al.(2020)Haghifam, Negrea, Khisti, Roy, and
  Dziugaite}]{haghifam2020sharpened}
Haghifam, M.; Negrea, J.; Khisti, A.; Roy, D.~M.; and Dziugaite, G.~K. 2020.
\newblock Sharpened generalization bounds based on conditional mutual
  information and an application to noisy, iterative algorithms.
\newblock \emph{Advances in Neural Information Processing Systems}, 33:
  9925--9935.

\bibitem[{Harsanyi(1963)}]{harsanyi}
Harsanyi, J.~C. 1963.
\newblock A simplified bargaining model for the n-person cooperative game.
\newblock \emph{International Economic Review}, 4(2): 194--220.

\bibitem[{He et~al.(2016)He, Zhang, Ren, and Sun}]{resnet}
He, K.; Zhang, X.; Ren, S.; and Sun, J. 2016.
\newblock Deep residual learning for image recognition.
\newblock In \emph{Proceedings of the IEEE conference on computer vision and
  pattern recognition}, 770--778.

\bibitem[{Keskar et~al.(2016)Keskar, Mudigere, Nocedal, Smelyanskiy, and
  Tang}]{flat_minima}
Keskar, N.~S.; Mudigere, D.; Nocedal, J.; Smelyanskiy, M.; and Tang, P. T.~P.
  2016.
\newblock On large-batch training for deep learning: Generalization gap and
  sharp minima.
\newblock \emph{arXiv preprint arXiv:1609.04836}.

\bibitem[{Krizhevsky, Hinton et~al.(2009)}]{cifar10}
Krizhevsky, A.; Hinton, G.; et~al. 2009.
\newblock Learning multiple layers of features from tiny images.

\bibitem[{Krizhevsky, Sutskever, and Hinton(2017)}]{alexnet}
Krizhevsky, A.; Sutskever, I.; and Hinton, G.~E. 2017.
\newblock Imagenet classification with deep convolutional neural networks.
\newblock \emph{Communications of the ACM}, 60(6): 84--90.

\bibitem[{Kwon et~al.(2021)Kwon, Kim, Park, and Choi}]{kwon2021asam}
Kwon, J.; Kim, J.; Park, H.; and Choi, I.~K. 2021.
\newblock Asam: Adaptive sharpness-aware minimization for scale-invariant
  learning of deep neural networks.
\newblock In \emph{International Conference on Machine Learning}, 5905--5914.
  PMLR.

\bibitem[{LeCun et~al.(1998)LeCun, Bottou, Bengio, and Haffner}]{mnist}
LeCun, Y.; Bottou, L.; Bengio, Y.; and Haffner, P. 1998.
\newblock Gradient-based learning applied to document recognition.
\newblock \emph{Proceedings of the IEEE}, 86(11): 2278--2324.

\bibitem[{Li et~al.(2018)Li, Xu, Taylor, Studer, and Goldstein}]{loss_lanscape}
Li, H.; Xu, Z.; Taylor, G.; Studer, C.; and Goldstein, T. 2018.
\newblock Visualizing the loss landscape of neural nets.
\newblock \emph{Advances in neural information processing systems}, 31.

\bibitem[{Li and Zhang(2023)}]{li2023does}
Li, M.; and Zhang, Q. 2023.
\newblock Does a Neural Network Really Encode Symbolic Concept?
\newblock In \emph{International Conference on Machine Learning}.

\bibitem[{Madry et~al.(2018)Madry, Makelov, Schmidt, Tsipras, and
  Vladu}]{adversarial_attacks}
Madry, A.; Makelov, A.; Schmidt, L.; Tsipras, D.; and Vladu, A. 2018.
\newblock Towards Deep Learning Models Resistant to Adversarial Attacks.
\newblock In \emph{International Conference on Learning Representations}.

\bibitem[{Neyshabur et~al.(2017)Neyshabur, Bhojanapalli, McAllester, and
  Srebro}]{generalization_bounds}
Neyshabur, B.; Bhojanapalli, S.; McAllester, D.; and Srebro, N. 2017.
\newblock Exploring generalization in deep learning.
\newblock \emph{Advances in neural information processing systems}, 30.

\bibitem[{Neyshabur, Tomioka, and Srebro(2015)}]{norm_bound}
Neyshabur, B.; Tomioka, R.; and Srebro, N. 2015.
\newblock Norm-based capacity control in neural networks.
\newblock In \emph{Conference on Learning Theory}, 1376--1401. PMLR.

\bibitem[{Ren et~al.(2023)Ren, Li, Chen, Deng, and Zhang}]{ren2021AOG}
Ren, J.; Li, M.; Chen, Q.; Deng, H.; and Zhang, Q. 2023.
\newblock Defining and Quantifying the Emergence of Sparse Concepts in DNNs.
\newblock \emph{Proceedings of the IEEE/CVF conference on computer vision and
  pattern recognition}.

\bibitem[{Ren et~al.(2024)Ren, Gao, Shen, and Zhang}]{ren2023arrived}
Ren, Q.; Gao, J.; Shen, W.; and Zhang, Q. 2024.
\newblock Where We Have Arrived in Proving the Emergence of Sparse Symbolic
  Concepts in AI Models.
\newblock \emph{International Conference on Learning Representations}.

\bibitem[{Simonyan and Zisserman(2014)}]{vgg}
Simonyan, K.; and Zisserman, A. 2014.
\newblock Very deep convolutional networks for large-scale image recognition.
\newblock \emph{arXiv preprint arXiv:1409.1556}.

\bibitem[{Weng et~al.(2018)Weng, Zhang, Chen, Yi, Su, Gao, Hsieh, and
  Daniel}]{CLEVER}
Weng, T.-W.; Zhang, H.; Chen, P.-Y.; Yi, J.; Su, D.; Gao, Y.; Hsieh, C.-J.; and
  Daniel, L. 2018.
\newblock Evaluating the robustness of neural networks: An extreme value theory
  approach.
\newblock \emph{arXiv preprint arXiv:1801.10578}.

\end{thebibliography}

\end{document}